# Cost-Sensitive Convolution based Neural Networks for Imbalanced Time-Series Classification

Yue Geng* and Xinyu Luo
*Mechanical and Electrical Engineering Institute of CUMTB, Beijing, 100083, China*
*E-mail: danielgy19890310@gmail.com*

**Abstract.** Some deep convolutional neural networks were proposed for time-series classification and class imbalanced problems. However, those models performed degraded and even failed to recognize the minority class of an imbalanced temporal sequences dataset. Minority samples would bring troubles for temporal deep learning classifiers due to the equal treatments of majority and minority class. Until recently, there were few works applying deep learning on imbalanced time-series classification (ITSC) tasks. Here, this paper aimed at tackling ITSC problems with deep learning. An adaptive cost-sensitive learning strategy was proposed to modify temporal deep learning models. Through the proposed strategy, classifiers could automatically assign misclassification penalties to each class. In the experimental section, the proposed method was utilized to modify five neural networks. They were evaluated on a large volume, real-life and imbalanced time-series dataset with six metrics. Each single network was also tested alone and combined with several mainstream data samplers. Experimental results illustrated that the proposed cost-sensitive modified networks worked well on ITSC tasks. Compared to other methods, the cost-sensitive convolution neural network and residual network won out in the terms of all metrics. Consequently, the proposed cost-sensitive learning strategy can be used to modify deep learning classifiers from cost-insensitive to cost-sensitive. Those cost-sensitive convolutional networks can be effectively applied to address ITSC issues.

## Introduction

Class imbalance problems (CIP) and time-series classification (TSC) are two top ten challenges in data mining [1].They have attracted increasing research enthusiasms from different communities over the past years. However, most previous works only focuses on addressing CIP and TSC issues separately. In fact, the combination (known as imbalanced time-series classification, ITSC) of TSC and CIP [2] could be frequently found in widespread real-life scenarios, such as behavior detection [3], medical treatments [4, 5], sleep monitoring [6], and industrial hazards surveillance [7-9]. Undoubtedly, ITSC is much more burdensome than the common TSC and CIP issues. This paper mainly focuses on binary ITSC problems.

ITSC is a special situation of TSC where one category is overrepresented compared to another class [2, 10]. In one unevenly distributed time-series dataset, majority class and minority class are referred as negative and positive respectively. Correct detection of positive class is difficult because it might be bound up with abnormal and significant cases [11]. It is important to learn a higher classification rate on rare events, meanwhile it is tolerable to ignore some majority instances misclassification [12]. For example, in continuous industrial surveillance, hazards warning activities occur as positive class which need more attention than normal ones. If a hazard activity is misdiagnosed as a normal one, the best rescue time ones. If a hazard activity is misdiagnosed as a normal one, the best rescue time would be delayed and it might cause serious consequences [8]. Thus, the research on how to correctly classify imbalanced temporal sequences data is significant.

Precious works can be summarized as two levels. In data manipulation level [2, 10, 13-15], time-series datasets were re-established through over-sampling of positive samples, or under-sampling of negative samples, or both. In algorithmic modification level [16], classifiers were modified by predefining higher costs or class weights for false positive samples. However, there were some problems of those two levels approaches needed to be noticed. Data preprocessing would change the raw data distribution in incremental or detrimental ways. Time-consuming problem and information loss are their main drawbacks. Algorithmic modification approaches need predefining the cost weight or cost matrix and the exact settings are difficultly found. Besides, most of two levels methods applied algorithms like KNN-DTM [17], SVM [13, 15], Shaplets [16]. Those classic algorithms need heavy hand-crafted works on data preprocessing or feature engineering and they are not appropriate for large volume dataset.

Recently, in the realms of TSC and CIP, some efforts were spent on exploiting deep neural networks for end-to-end classification. Unlike traditional TSC algorithms, deep learning approaches could capture time shift properties and invariant features of temporal sequences automatically [12, 17-23]. Meanwhile, convolution neural networks (CNNs) were applied on data sampling level [24-27] or algorithmic modification level [28, 29] to tackle imbalanced data issues. A systematic investigation can be found in [30]. However, most existing deep learning research just focused on common TSC and CIP cases. Until recently, there is some lack of knowledge about using deep learning classifiers to deal with ITSC problems.

This paper aimed to address binary ITSC issues with four CNN based models from [20, 22, 31] and a temporal multilayer perceptron (MLP). An adaptive cost-sensitive learning strategy was proposed to modify those networks from cost-insensitive to cost-sensitive. This ideal was inspired by [12, 28, 32] but different with them. The proposed method defined an adaptive updating misclassification penalty parameter. It was inserted into the loss function of modified classifier and optimized by global and local imbalanced ratios. Subsequently, those modified networks were validated on a real-life dataset (comes from a coal mine seismic monitoring system). In order to present unbiased results, this paper also conducted experiments on some mainstream re-sampling methods with the above CNN based structures.

This paper made a contribution to the study of ITSC with deep learning. However, more validations of the proposed method need to be done on some imbalanced time-series benchmark dataset and multi-class cases need to be considered in the next study or elsewhere. The remainder of this paper is organized as follows. Brief reviews of related works are given in Section 1. Section 2 describes four CNN based networks and the proposed cost-sensitive learning strategy. Section 3 presents the evaluation measures, experimental setting and results. Section 4 provides the discussion. Section 5 summarizes and concludes this paper.

# 1. Related works

Since the high dimensionality, under-representation and low proportion, positive time-series samples cannot be recognized easily. Thus, when facing ITSC cases, deep learning classifiers would perform degradedly. How to effectively address the above problem is a crucial issue. But there was little information available in literature about ITSC problems with deep learning. Therefore, this section gives briefly reviews of two relevant works.

*1.1. Imbalanced data classification via cost-sensitive deep learning*

Over-sampling and under-sampling are two classic data manipulation methods on tackling CIP issues. These techniques aim at changing the class distribution in preprocessing step and they are independent of the underlying classification models [11]. A range of certain achievements have been obtained by applying data sampling methods, like random sampling and some synthetic sampling [33]. However, there are some problems worth noting. Data sampling techniques change the distribution of raw data and would bring some drawbacks. In particular, over-sampling may aggravate the burden of computation and cause overfitting; while under-sampling may lose some useful information [28].

Algorithmic level approaches aim at modifying the insensitive models into minority sensitive ones. Cost-sensitive enhances the classifier performance towards CIP by penalizing each type of misclassification error differently according to predefined costs. Matjaz Kukar and Igor Kononenko [34] reported a fundamental cost-sensitive modification on one multilayered feedforward network. They changed the learning rate giving higher cost instances higher punishing weights and misclassification cost objective function was used. Zhi-Hua Zhou and Xu-Ying Liu [23] empirically investigated a cost-sensitive neural network. They also discussed the effects of sampling and threshold-moving on the training stages. The above works modified neural networks into cost-sensitive classifiers by fixing cost-matrix. However, the manual setting cost matrix relied on professional judgements and this might be constrained in some applications.

Recently, some cost-sensitive deep neural networks were introduced to the CIP domain. S. H. Khan et al. [28] proposed a cost-sensitive deep neural network with automatic feature represented. They synchronously optimized the parameters of CNNs with learnable cost parameters to execute the cost sensitive operations. Vidwath Ral et al. [12] explored cost sensitive CNNs with different cost functions and provided competitive results. Shoujin Wang et al. [35] defined two loss functions named mean false error and mean squared false error. They were applied to make deep neural networks more sensitive to the minority class, aiming for higher overall accuracy. Mateusx Buda et al. [30] investigated the impactes of CIP on CNNs and validated them on three graphic datasets. They also gave a conclusion that the drawbacks of sampling may not degrade CNN classification performance. Nevertheless, their statement was observed from graphic experiments and it was uncertain for imbalanced time-series datasets. Based on the above solid works, this paper constructed four cost-sensitive CNN based classifiers with the proposed method. Particularly, some popular convolutional structures like basic CNN, fully convolutional network (FCN), residual network (ResNet) and long short-term memory FCN (LSTM-FCN) were built in Section 2. In addition, this paper also investigated the conclusion in [30] on one real-life dataset with skewed class distribution.

*1.2. Time-series classification via deep convolutional neural networks*

Some deep neural networks were applied on avoiding the drawbacks of conventional time-series classifiers. While on the first glimpse, it seems that recurrent neural networks (like long short-term memory, LSTM) could match TSC tasks naturally. But some works have proved that CNNs perform well or even better [7]. Yi Zheng et al. [22, 36] proposed a novel deep learning model named multi-channels CNN for multivariate TSC. They automatically extracted temporal features by one-dimensional multi-channels convolution layers and classifier the abstracted features by a fully connected perceptron. Soon afterwards, Zhicheng Cui et al. [18] improved the performance of CNN on TSC by transforming raw time-series in a multiscale way. They called the method as MCNN and validated it on several univariate datasets. Similar with MCNN, Wenlin Wang et al. [21] presented an Earliness-Aware Deep Convolutional Network (EA-ConvNet) for early classification of time-series. They claimed that the traditional feature based TSC algorithm shaplets was a special case of features learned by their method.

Above works applied basic CNNs, but some popular extended structures have been shown to achieve competitive performance on TSC tasks. Zhiguang Wang et al. [31] explored FCN and ResNet on addressing TSC problems. LSTMs could able to capture time-ordered dependencies in time-series. Considering that, Fazle Karim et al. [20] parallel connected LSTM with FCN to recognize hidden patterns in sequences. Both of Zhiguang Wang and Fazle Karim used the University of California Riverside (UCR) Benchmark datasets [37]. Some convincing experimental results were given in their works. The main advantage of CNN based solutions is that they do not require heavy preprocessing or hand-craft feature engineering. But they were just designed to address the normal TSC problems without considering the CIP issues. Consequently, their performance cannot be guaranteed on imbalanced time-series datasets. Thus, this paper aimed to take advantage of those CNN based models and furthermore modify them into ITSC effective algorithms.

## 2. Methods

In this section, a cost-sensitive learning strategy is proposed to execute the ITSC tasks. It can be used to modify deep learning models from cost-insensitive into cost-sensitive.

*2.1. Preliminaries*

According to [18, 38], a time-series can be defied as a time ordered real-values $\mathbf{T} = \{t_1, t_2, \ldots, t_l\}$, where $l$ is the length of $\mathbf{T}$. While a multivariate time series is a vector of time series $\mathbf{S} = (\mathbf{T}_1, \mathbf{T}_2, \ldots, \mathbf{T}_L)$, where each element represents a temporal sequence. Thus, a supervised temporal sequence can be defined as $\mathbf{D} = \{(\mathbf{S}_i, Y_i)\}_{i=1}^n$, where $\mathbf{S}_i$ and $Y_i$ represent the $i^{th}$ multivariate temporal sequences and the relevant label respectively. This paper mainly focused on binary time-series classification and the relevant labels were denoted as $Y \in \{0, 1\}$.

## 2.2. Temporal convolutional neural networks

In [36], the authors used one-dimensional convolutional layers to automatically extract features from multivariate time-series raw data. They applied this mechanism to modify standard CNNs to execute TSC tasks. A similar structure was applied in this paper, which included convolutional and pooling layers, rectified linear units (ReLU) [39] and dropout operations. Three one-dimensional filters were operated as feature extractors and one fully connected layer performed the final binary classification. The details of this architecture was illustrated in **Figure 1**. The numbers in grey shadow blocks indicate the one-dimensional convolutional layer size. The three parameters illustrate the number of convolution kernels, the dimension size and sliding window size respectively. The dash line means dropout operation.

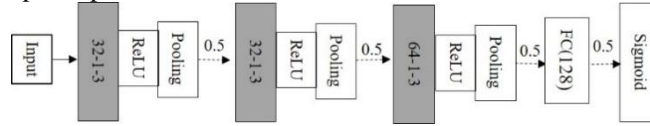

**Figure 1.** Multi-channels temporal CNN structure

Except for temporal CNN, some other deep learning models were modified by applying one-dimensional temporal layers. Like temporal FCN [31], temporal ResNet [31] and temporal LSTM-FCN [20]. Their structures were shown in figures from **Figure 2** to **4**. However, the above temporal deep learning models were designed without considering the class imbalanced problems.

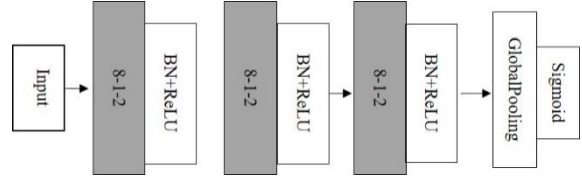

**Figure 2.** Multi-channels temporal FCN structure

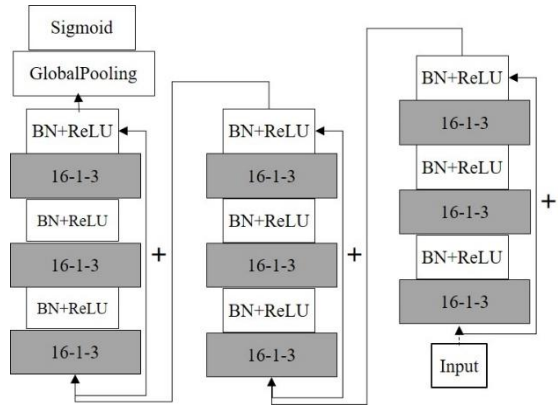

**Figure 3.** Multi-channels temporal ResNet structure

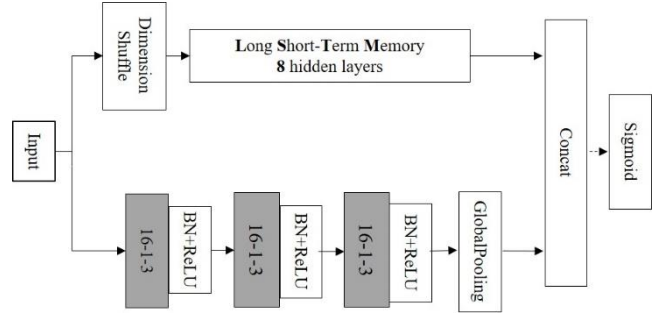

**Figure 4.** LSTM-FCN: hybrid structure of temporal FCN combined with LSTM

*2.3. Cost-sensitive*

The major advantage of cost-sensitive classifiers is the distinction of treatments on majority and minority classes. They also consider misclassification costs. According to [40], the misclassification costs can be presented as a confusion matrix in Table 1. 0 and 1 represent negative (majority) class and positive (minority) class respectively. The rows are actual classes and the columns are predicted ones. TP and TN are the correctly classified positive and negative samples respectively, while FP and FN are the misclassified positive and negative samples respectively. The geometric value and accuracy rate can be derived from confusion matrix as Eq. (1) and (2).

$$G_{mean} = \sqrt{\frac{TP}{TP+FN} \times \frac{FP}{TN+FP}} \quad (1)$$

$$ACC = \frac{TP+TN}{TP+TN+FP+FN} \quad (2)$$

Table 1 Binary Classification Confusion Matrix

|  | Actual Negative | Actual Positive |
| --- | --- | --- |
| Predict Negative | True Negative(**TN**) | False Negative(**FN**) |
| Predict Positive | False Positive(**FP**) | True Positive(**TN**) |

According to the minimum expected cost principle, the expected risk of one cost-sensitive classifier is formulized as Eq.(3).

$$R(i|\mathbf{S}) = \sum_{j} P(j|\mathbf{S})C(j,i) \quad (3)$$

where $R(i|\mathbf{S})$ is the expected risk of categorizing the given input $\mathbf{S}$ into class $i$. $P(j|\mathbf{S})$ is the posterior probability when the given input $\mathbf{S}$ belong to class $j$. $C(j,i)$ is the misclassification cost when class $i$ is misrecognized as class $j$.

However, it is always difficult to calculate the accurate posterior probability. Thus, some works [41] applied the empirical risk instead of seeking the posterior probability in neural networks as Eq. (4) and (5).

$$\hat{R}_l(o) = \mathbb{E}_{\mathbf{S},\mathbf{Y}}[l] = \frac{1}{n}\sum_{i=1}^{n} l(C, d^{(i)}, o^{(i)}) \quad (4)$$

$$C = \begin{cases} C_{p,q} = 1, p = q \\ C_{p,q} = IR, p \neq q \end{cases} \quad (5)$$

where $\hat{R}_l(o)$ is the empirical risk. $\mathbb{E}$ is the minimum expectation. **S** represents the given time series training samples. **Y** represents the relevant class labels. $n$ is the overall number of multivariate time series instances. $C$ is the cost matrix in which the cost will be set as imbalance ratio $IR$ when predicted class $q$ matches the actual class $p$. $o^{(i)}$ and $d^{(i)}$ represents the $i^{th}$ predicted output and desired output respectively. $l(\cdot)$ represents the loss function of the network.

*2.4. The proposed cost-sensitive learning strategy*

Applying imbalanced ratio as the penalty on cost-sensitive misclassification may alleviate the class imbalance problem in an overall view. However, fixed cost matrix could not fit imbalanced distribution of local areas like minibatched training sets of CNNs. Thus, the proposed method utilized a dynamically changing misclassification cost weight which could be adaptively updated. It was based on the imbalanced distributions of not only the whole training set but also local minibathches. For dealing with ITSC problems, those convolutional classifiers mentioned in Section 2.2 were modified with the cost-sensitive learning strategy.

The cross entropy loss of $n^{th}$ training instance can be expressed by

$$LOSS(\boldsymbol{\theta}) = \lambda \times d_n \times (-\ln(y_n)) + (1 - d_n) \times (-\ln(1 - y_n)) \quad (6)$$

where $\boldsymbol{\theta}$ is the weight parameters of applied classifier (like CNN). $\lambda$ is the proposed misclassification cost. $d_n$ and $y_n$ are the desired output and predicted output respectively in $n^{th}$ training instance.

Therefore, the overall cross entropy loss can be computed by adding the pos class error and the neg class error. The overall loss function and optimization are shown as Eq. (7) to (9).

$$E(\boldsymbol{\theta}) = \frac{1}{n^{pos}} \sum_{i=1}^{n^{pos}} LOSS^{pos}(\boldsymbol{\theta}^{pos}, \lambda_n^{pos}) + \frac{1}{n^{neg}} \sum_{i=1}^{n^{neg}} LOSS^{neg}(\boldsymbol{\theta}^{neg}, \lambda_n^{neg}) \quad (7)$$

$$\lambda_n = \begin{cases} IR^{overall} \times \exp(-\frac{G_{mean}^{batch}}{2}) \times \exp(-\frac{Acc^{batch}}{2}), if \quad n \in neg \\ 1, if \quad n \in pos \end{cases} \quad (8)$$

$$(\boldsymbol{\theta}^*) = \arg\min E(\boldsymbol{\theta}) \quad (9)$$

where $IR^{overall}$ is the overall imbalance ratio. $G_{mean}^{batch}$ and $Acc^{batch}$ are geometric value and accuracy of the current minibatch training examples respectively.

In particular, the local metrics ($G_{mean}^{batch}$ and $Acc^{batch}$) would be updated after each minibacth and consequently the proposed $\lambda$ would be updated in an adaptive manner. Specifically, a random shuffle trick was applied on the input training imbalanced temporal sequences sets and minibatches were assigned from those shuffled time series samples. This trick was aim at avoiding the absence of minority samples in each minibatch and improve the generalization of the applied classifiers.

According to the statement of [12, 28], cost matrix would not affect gradient decent processing and only change binary output. Similarly, the proposed misclassification cost weight would not bother the optimization processes either. It only forced the above CNN based networks into cost-sensitive and class imbalance effective models. The proposed cost-sensitive approach can be summarized as Algorithm 1.

**Algorithm 1** Optimization for parameter $\theta$ of Cost-sensitive CNN

**Input:** Imbalanced time-series set, Maximum epoch: $M$, Batch Size: $B$, learning rate: $\eta$
**Output:** $\theta^*$
1: Imbalanced time-series set, Maximum epoch: $M$, Batch Size: $B$, learning rate: $\eta$
2: Randomly initialize weight $\theta$
3: Calculate the overall imbalanced ratio ($IR$)
4: Random shuffle the imbalanced time series training set and assign minibatches
5: **for** $[1,M]$ **do**
6:    **for** $[1,B]$ **do**
7:       forward passing
8:       Calculate $G_{mean}$ and ACC of the current minibatch as Eq.(1) and Eq.(2)
9:       Update the misclassification cost weight $\lambda$ as Eq.(4)
10:      Calculate loss as Eq.(3)
11:      Gradients Calculation
12:      Update $\theta$
13: **return** $\theta^*$

## 3. Experiments and Results

### 3.1. Performance Evaluation Metrics for CIP

Evaluation metrics choosing would affect the objectivity and fairness of final assessments. In a traditional way, most classifiers were empirically assessed by overall accuracy rate. However, since it cannot reflect false positive samples, ACC is not appropriate any more when facing CIP. Therefore, it is necessary to alternate the overall accuracy with some more effective metrics. Some performance evaluation metrics were picked to assess classification performance of the applied classifiers.

(1) True positive rate (TPR) is also called recall or sensitive, which reflects the correct classification proportions of positive samples.

$$TPR = \frac{TP}{TP + FN} \quad (10)$$

(2) True negative rate (TNR) is also called specificity, which reflects the correct classification proportions of negative samples.

$$TNR = \frac{TN}{FP + TN} \quad (11)$$

(3) False positive rate (FPR) reflects the misclassification proportions of positive samples.

$$FPR = \frac{FP}{FP + TN} \quad (12)$$

(4) Positive predictive value (PPV) is also called precision, which reflects the correct predicted proportions of all positive samples.

$$PPV = \frac{TP}{TP + FP} \quad (13)$$

(5) $F_1$ score is the harmonic mean of precision and sensitive, which means that precision is as important as sensitive.

$$F_1 = \frac{2PPV \cdot TPR}{PPV + TPR} \quad (14)$$

(6) $G_{mean}$ is the geometric mean of TNR and TPR.

$$G_{mean} = \sqrt{TPR \cdot TNR} \quad (15)$$

(7) Receiver operating characteristic curve (ROC Curve) is plotting TPR against FPR, while precision recall curve (PR Curve) is plotting precision against recall. Usually, they are measured by the area under the curve (AUC). ROCAUC and PRAUC are applied in this paper.

*3.2. Experimental Dataset*

The experimental dataset was obtained from AAIA16 Data Mining Challenge: Predicting Dangerous Seismic Events in Active Coal Mines [8]. The organizer Knowledge Pit[1] offered one large volume dataset to predict increased seismic activities that endanger coal workers working underground. The monitored seismic dataset was separated into training set and testing set, more details shown in Table 2. In training part, five separated individual training sets were given and their combination contained 541 attributes and 133,151 samples. Class labels were given in normal and warning class. The training set was collected via more than five years. The test part consists of 3,860 examples and covers a period of nearly 16 months. Each data sample is described by a series of hourly aggregated sensor readings from one whole day period. The class distribution of this dataset is imbalanced, there are 130,188 normal samples and 2963 warning samples in training set. 3664 normal samples and 196 warning samples are included in test set. This imbalanced time-series dataset was used to validate the proposed cost-sensitive CNN based classifiers.

Table 2 Coal Mine Seismic Dataset

|  | Number of Samples | Pos | Neg | IR |
|---|---|---|---|---|
| Training | 133151 | 2963(2.225%) | 130188(97.775%) | 43.9379 |
| Testing | 3860 | 196(5.078%) | 3664(94.922%) | 18.6939 |

*3.3. Experiment Setup*

There were four temporal CNN based models (see **Figures** from **1** to **4**) involving in this section that mentioned in Section 2, they were temporal CNN, temporal FCN, temporal ResNet and temporal LSTM-FCN. In addition, a three layered temporal MLP was also implemented to compare with the four CNN based models. For tackling the ITSC issues, three experiments were conducted. (1) Imbalanced time-series classification with five single networks. (2) Preprocessing the seismic dataset via fifteen mainstream data sampling methods and classifying the sampled temporal sequences with five networks. (3) Modifying those networks with the proposed cost-sensitive strategy and classifying imbalanced seismic multivariate sequences with modified networks.

---

[1] A polish data challenge platform: https://knowledgepit.fedcsis.org/

Except normalization and one-hot encoding, the proposed method did not commit any manual feature engineering operation. The structures of four based networks are illustrated in **Figures** from **1** to **4**. There were two 32-neuron layers and one 64-neuron layer in the applied MLP. All activation functions of hidden layers were applied with rectified linear units (ReLU) and the final output layers were sigmoid units. All networks were trained with Adam [31] in which the learning rate is 0.001, $\beta_1 = 0.9$, $\beta_2 = 0.999$ and $\varepsilon = 1e-8$. Cross entropy loss functions were changed by inserting adaptive misclassification penalties. Applied data sampling approaches included over-sampling, under-sampling and two techniques combined methods, as it is shown in Table 3. For comprehensive and objective evaluating, this work adopted 10-fold cross validation to measure Recall, Precision, $F_1$, $G_{mean}$, ROCAUC and PRAUC. Batch size was fixed as 512 in this work. Those neural networks were implemented in deep learning framework TensorFlow and speeded up on a GTX1080 GPU. Data sampling methods were imported from a python package imbalanced-learn [42]. All experiments were executed on a PC with one Intel i7-6700K 4.0GHz processor and 32GB of RAM.

Table 3 Mainstream Data Sampling Methods

| Over-sample | Under-sampling | Combined Sampling |
| --- | --- | --- |
| Random Over Sampling (ROS) | Random Under Sampling (RUS) | SMOTE+ENN |
| SMOTE | InstanceHardnessThreshold (IHT) | SMOTE+ TL |
| SMOTE Borderline 1 (b1) | NearMiss (NM) | |
| SMOTE Borderline 2 (b2) | TomekLinks (TL) | |
| SMOTE SVM | EditedNearestNeighbours (ENN) | |
| ADASYN | OneSidedSelection (OSS) | |
| | NeighbourhoodCleaningRule (NCR) | |

*3.4. Imbalanced time-series classification with single neural networks*

Precious studies have given evidence that temporal CNNs could handle TSC problems effectively. However, when facing CIP, the performance of those networks need to be investigated. Hence, five single networks were tested on one real-life coal mine seismic dataset. Their performance was summarized in Table 4, in which mean values are given and standard deviation is shown in parentheses. The best results on each metrics were emphasized in bold-face. **Figure 5** illustrates the evaluation metrics comparison of five single network classifiers.

The stipulated score measure of AAIA'16 was ROCAUC. If only using ROCAUC metric in this work, all the models performed well and CNN ranked first. But when all the metrics were taken into account, the performance of those single classifiers was not satisfied. As it is shown in Table 4 and **Figure 5**, Precision and $F_1$ metrics are terribly measured. Additionally, zero valued metrics of MLP, FCN and ResNet are plotted by absent bars in **Figure 5**. Furthermore, there are several invalid measures (represented by nan) in columns Recall and $G_{mean}$. When TP and FN are zeros, Recall would by divided by zeros and this would cause invalid results. It seems that the applied classifiers totally fail.

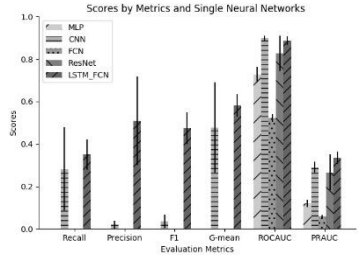

**Figure 5.** Metrics Comparison of Five Single Networks.

Table 4 Evaluation results of five pure networks

| Networks | Recall | Precision | $F_1$ | $G_{mean}$ | ROCAUC | PRAUC |
|---|---|---|---|---|---|---|
| MLP | nan | 0.0000 | 0.0000 | nan | 0.7265 | 0.1205 |
|  | nan | (0.0000) | (0.0000) | nan | (0.0357) | (0.0171) |
| MDCNN | 0.2828 | 0.0194 | 0.0359 | 0.4778 | **0.8983** | 0.2907 |
|  | (0.1962) | (0.0177) | (0.0311) | (0.2120) | (0.0117) | (0.0262) |
| FCN | nan | 0.0000 | 0.0000 | nan | 0.5226 | 0.0587 |
|  | nan | (0.0000) | (0.0000) | nan | (0.0187) | (0.0061) |
| ResNet | nan | 0.0000 | 0.0000 | nan | 0.8277 | 0.2666 |
|  | nan | (0.0000) | (0.0000) | nan | (0.0829) | (0.0845) |
| LSTM-FCN | **0.3503** | **0.5087** | **0.4742** | **0.5811** | 0.8865 | **0.3345** |
|  | (0.0713) | (0.2078) | (0.0750) | (0.0540) | (0.0205) | (0.0292) |

*3.5. Imbalanced time-series classification with data sampled neural networks*

Data sampling re-established the skewed dataset through over-sampling, or under-sampling or two techniques combined. Several mainstream data samplers (seeing as Table 3) were picked to work with those five single networks together. Averages of all data sampling percentages are illustrated in Tables 5 through 9. The best results on each metrics are emphasized in bold-face, and standard deviation is shown in parentheses. The average performance of each neural network on applying those fifteen data sampling approaches is plotted in **Figure 6**.

Obviously, samplers combined with single networks perform better than only applying single networks, especially in the term of Precision. But there are still some problems like invalidations of Recall and $G_{mean}$ measures, zero Precision measures. Performance measures of different sampling methods are located in a wide range. Experimental results illustrate that data sampling might improve the performance of single networks, but affections of different data manipulation methods vary. The average performance is plotted in **Figure 6**. Data sampled LSTM-FCN classifier performs best. Second ranking could be data sampled ResNet, but it is exceeded by data sampled CNN in the term of Precision.

Table 5 Evaluation results of data sampling combined with MLP

| Methods | Recall | Precision | $F_1$ | $G_{mean}$ | ROCAUC | PRAUC |
|---|---|---|---|---|---|---|
| ROS | 0.0918 | 0.7500 | 0.1771 | 0.2986 | 0.7456 | 0.1505 |
|  | (0.0155) | (0.0995) | (0.0278) | (0.0252) | (0.0253) | (0.0219) |
| SMOTE | 0.0916 | 0.7388 | 0.1769 | 0.2981 | 0.7384 | 0.1476 |
|  | (0.0160) | (0.1009) | (0.0297) | (0.0270) | (0.0459) | (0.0430) |
| SMOTE b1 | 0.1016 | 0.6617 | 0.1877 | 0.3119 | 0.7373 | 0.1485 |
|  | (0.0281) | (0.2012) | (0.0499) | (0.0438) | (0.0539) | (0.0368) |
| SMOTE b2 | 0.1001 | 0.6811 | 0.1853 | 0.3090 | 0.7383 | 0.1464 |
|  | (0.0316) | (0.1684) | (0.0461) | (0.0470) | (0.0424) | (0.0327) |
| SMOTE SVM | 0.1133 | 0.5837 | **0.1958** | 0.3289 | 0.7409 | 0.1695 |
|  | (0.0341) | (0.2402) | (0.0463) | (0.0461) | (0.0514) | (0.0377) |
| ADASYN | 0.0966 | 0.6597 | 0.1827 | 0.3046 | 0.7126 | 0.1269 |
|  | (0.0233) | (0.0748) | (0.0402) | (0.0360) | (0.0363) | (0.0334) |
| RUS | 0.0713 | 0.8643 | 0.1406 | 0.2637 | 0.7205 | 0.1411 |
|  | (0.0123) | (0.0765) | (0.0231) | (0.0223) | (0.0456) | (0.0351) |
| IHT | **0.2756** | 0.0321 | 0.0562 | **0.5014** | 0.7784 | 0.1538 |
|  | (0.1249) | (0.0222) | (0.0366) | (0.1093) | (0.0378) | (0.0255) |
| NM | 0.0566 | **0.9245** | 0.1126 | 0.2350 | 0.6156 | 0.0964 |
|  | (0.0047) | (0.0591) | (0.0088) | (0.0096) | (0.0187) | (0.0113) |
| TL | nan | 0.0000 | 0.0000 | nan | 0.7575 | 0.1307 |
|  | nan | (0.0000) | (0.0000) | nan | (0.0245) | (0.0196) |
| ENN | nan | 0.0000 | 0.0000 | nan | 0.7490 | 0.1430 |
|  | nan | (0.0000) | (0.0000) | nan | (0.0400) | (0.0232) |
| OSS | nan | 0.0000 | 0.0000 | nan | 0.7700 | 0.1572 |
|  | nan | (0.0000) | (0.0000) | nan | (0.0288) | (0.0219) |
| NCR | nan | 0.0000 | 0.0000 | nan | 0.7447 | 0.1286 |
|  | nan | (0.0000) | (0.0000) | nan | (0.0253) | (0.0210) |
| SMOTE+ENN | 0.0926 | 0.8378 | 0.1797 | 0.2980 | 0.7719 | **0.1824** |
|  | (0.0327) | (0.1067) | (0.0589) | (0.0499) | (0.0395) | (0.0262) |
| SMOTE+TL | 0.0581 | 0.8327 | 0.1065 | 0.2393 | 0.6781 | 0.1159 |
|  | (0.0093) | (0.3014) | (0.0175) | (0.0173) | (0.0564) | (0.0438) |
| Average | 0.1045 | 0.5044 | 0.1134 | 0.3081 | 0.7333 | 0.1426 |
| Performance | (0.0302) | (0.0967) | (0.0257) | (0.0394) | (0.0381) | (0.0289) |

Table 6 Evaluation results of data sampling combined with CNN

| Methods | Recall | Precision | $F_1$ | $G_{mean}$ | ROCAUC | PRAUC |
|---|---|---|---|---|---|---|
| ROS | 0.1472 | 0.9327 | 0.2912 | 0.3825 | 0.9005 | 0.3242 |
|  | (0.0095) | (0.0122) | (0.0187) | (0.0124) | (0.0071) | (0.0224) |
| SMOTE | 0.1402 | 0.9097 | 0.2765 | 0.3727 | 0.8779 | 0.3029 |
|  | (0.0150) | (0.0132) | (0.0290) | (0.0200) | (0.0129) | (0.0307) |
| SMOTE b1 | 0.1609 | 0.9082 | 0.3167 | 0.3996 | 0.8917 | 0.3282 |
|  | (0.0107) | (0.0152) | (0.0208) | (0.0134) | (0.0121) | (0.0327) |
| SMOTE b2 | 0.1144 | 0.8786 | 0.2252 | 0.3364 | 0.8148 | 0.1733 |
|  | (0.0046) | (0.0133) | (0.0093) | (0.0070) | (0.0178) | (0.0211) |
| SMOTE SVM | 0.1595 | 0.9214 | 0.3145 | 0.3976 | 0.8959 | 0.3370 |
|  | (0.0175) | (0.0130) | (0.0338) | (0.0220) | (0.0122) | (0.0230) |
| ADASYN | 0.0736 | 0.8582 | 0.1455 | 0.2689 | 0.6759 | 0.1467 |
|  | (0.0031) | (0.0255) | (0.0063) | (0.0061) | (0.0208) | (0.0217) |
| RUS | 0.1359 | 0.9000 | 0.2678 | 0.3671 | 0.8676 | 0.2359 |

|  | (0.0084) | (0.0172) | (0.0166) | (0.0113) | (0.0125) | (0.0252) |
|---|---|---|---|---|---|---|
| IHT | 0.3242 | 0.6179 | **0.5387** | 0.5632 | 0.9022 | 0.3200 |
|  | (0.0083) | (0.0683) | (0.0290) | (0.0074) | (0.0058) | (0.0151) |
| NM | 0.0485 | 0.8342 | 0.0961 | 0.2128 | 0.4760 | **0.3796** |
|  | (0.0035) | (0.0700) | (0.0073) | (0.0103) | (0.0381) | (0.0351) |
| TL | 0.3664 | 0.0878 | 0.1514 | 0.5852 | 0.8839 | 0.2964 |
|  | (0.1094) | (0.0332) | (0.0538) | (0.0873) | (0.0178) | (0.0266) |
| ENN | **0.4010** | 0.1883 | 0.2880 | **0.6173** | 0.8773 | 0.3065 |
|  | (0.0760) | (0.0383) | (0.0429) | (0.0574) | (0.0128) | (0.0243) |
| OSS | 0.3402 | 0.0612 | 0.1073 | 0.5669 | 0.9014 | 0.3098 |
|  | (0.0629) | (0.0314) | (0.0500) | (0.0526) | (0.0078) | (0.0124) |
| NCR | 0.3792 | 0.2020 | 0.2984 | 0.6026 | **0.9047** | 0.3306 |
|  | (0.0228) | (0.0574) | (0.0697) | (0.0179) | (0.0061) | (0.0133) |
| SMOTE+ENN | 0.1444 | **0.9347** | 0.2860 | 0.3789 | 0.8983 | 0.3329 |
|  | (0.0111) | (0.0075) | (0.0218) | (0.0143) | (0.0067) | (0.0198) |
| SMOTE+TL | **0.1481** | 0.9235 | 0.2925 | 0.3835 | 0.8894 | 0.3139 |
|  | (0.0086) | (0.0127) | (0.0166) | (0.0112) | (0.0064 ) | (0.0218) |
| Average | 0.2056 | 0.6772 | 0.2597 | 0.4290 | 0.8438 | 0.2959 |
| Performance | (0.0248) | (0.0286) | (0.0284) | (0.0234) | (0.0131) | (0.0230) |

Table 7 Evaluation results of data sampling combined with FCN

| Methods | Recall | Precision | $F_1$ | $G_{mean}$ | ROCAUC | PRAUC |
|---|---|---|---|---|---|---|
| ROS | 0.0701 | 0.7908 | 0.1180 | 0.2557 | 0.6811 | 0.1181 |
|  | (0.0326) | (0.2994) | (0.0193) | (0.0511) | (0.0508) | (0.0352) |
| SMOTE | 0.0633 | 0.9352 | 0.1238 | 0.2472 | 0.7629 | 0.1408 |
|  | (0.0241) | (0.1467) | (0.0399) | (0.0394) | (0.0434) | (0.0340) |
| SMOTE b1 | 0.0618 | **0.9439** | 0.1218 | 0.2445 | 0.7445 | 0.1279 |
|  | (0.0209) | (0.1107) | (0.0368) | (0.0351) | (0.0236) | (0.0312) |
| SMOTE b2 | 0.0593 | 0.8985 | 0.1174 | 0.2406 | 0.6908 | 0.1334 |
|  | (0.0084) | (0.1114) | (0.0152) | (0.0162) | (0.0562) | (0.0331) |
| SMOTE SVM | 0.0614 | 0.9357 | 0.1221 | 0.2460 | **0.7629** | **0.1660** |
|  | (0.0071) | (0.0850) | (0.0139) | (0.0138) | (0.0666) | (0.0476) |
| ADASYN | 0.0578 | 0.7990 | 0.1073 | 0.2360 | 0.5908 | 0.0788 |
|  | (0.0100) | (0.3026) | (0.0212) | (0.0183) | (0.0351) | (0.0117) |
| RUS | **0.0861** | 0.8577 | **0.1259** | **0.2719** | 0.7386 | 0.1552 |
|  | (0.0823) | (0.2749) | (0.0259) | (0.1005) | (0.090) | (0.0337) |
| IHT | nan | 0.0000 | 0.0000 | nan | 0.5259 | 0.0585 |
|  | nan | (0.0000) | (0.0000) | nan | (0.0190) | (0.0054) |
| NM | 0.0524 | 0.8786 | 0.1034 | 0.2204 | 0.5108 | 0.0545 |
|  | (0.0040) | (0.1794) | (0.0050) | (0.0131) | (0.0325) | (0.0069) |
| TL | nan | 0.0000 | 0.0000 | nan | 0.5351 | 0.0613 |
|  | nan | (0.0000) | (0.0000) | nan | (0.0147) | (0.0048) |
| ENN | nan | 0.0000 | 0.0000 | nan | 0.5293 | 0.0611 |
|  | nan | (0.0000) | (0.0000) | nan | (0.0152) | (0.0056) |
| OSS | nan | 0.0000 | 0.0000 | nan | 0.5223 | 0.0600 |
|  | nan | (0.0000) | (0.0000) | nan | (0.0105) | (0.0047) |
| NCR | nan | 0.0000 | 0.0000 | nan | 0.5173 | 0.0570 |
|  | nan | (0.0000) | (0.0000) | nan | (0.0146) | (0.0057) |
| SMOTE+ENN | 0.0624 | 0.9010 | 0.1216 | 0.2432 | 0.6654 | 0.1196 |
|  | (0.0191) | (0.1659) | (0.0296) | (0.0346) | (0.0926) | (0.0514) |
| SMOTE+TL | 0.0581 | 0.8327 | 0.1065 | 0.2393 | 0.6781 | 0.1159 |
| | (0.0093) | (0.3014) | (0.0175) | (0.0173) | (0.0564) | (0.0438) |
| Average | 0.0633 | 0.5849 | 0.0779 | 0.2445 | 0.6304 | 0.1005 |
| Performance | (0.0218) | (0.1318) | (0.0150) | (0.0339) | (0.0387) | (0.0237) |

Table 8 Evaluation results of data sampling combined with ResNet

| Methods | Recall | Precision | $F_1$ | $G_{mean}$ | ROCAUC | PRAUC |
|---|---|---|---|---|---|---|
| ROS | 0.1304 | 0.8418 | 0.2538 | 0.3567 | 0.8362 | 0.2230 |
| | (0.0313) | (0.1185) | (0.0611) | (0.0423) | (0.0536) | (0.0603) |
| SMOTE | 0.1345 | 0.7393 | 0.2565 | 0.3601 | 0.8040 | 0.2250 |
| | (0.0372) | (0.1094) | (0.0689) | (0.0515) | (0.0658) | (0.0580) |
| SMOTE b1 | 0.1479 | 0.8163 | **0.2862** | 0.3803 | 0.8465 | 0.2645 |
| | (0.0300) | (0.1029) | (0.0583) | (0.0394) | (0.0612) | (0.0507) |
| SMOTE b2 | 0.0901 | 0.7679 | 0.1754 | 0.2961 | 0.7411 | 0.1643 |
| | (0.0167) | (0.1407) | (0.0347) | (0.0274) | (0.0851) | (0.0584) |
| SMOTE SVM | 0.1239 | **0.8985** | 0.2438 | 0.3480 | 0.8597 | 0.2713 |
| | (0.0303) | (0.0580) | (0.0582) | (0.0441) | (0.0486) | (0.0512) |
| ADASYN | 0.0626 | 0.7847 | 0.1231 | 0.2464 | 0.6102 | 0.0749 |
| | (0.0036) | (0.0508) | (0.0074) | (0.0077) | (0.0576) | (0.0228) |
| RUS | 0.0912 | 0.7964 | 0.1785 | 0.2969 | 0.7478 | 0.1709 |
| | (0.036) | (0.1549) | (0.0474) | (0.0412) | (0.1162) | (0.0762) |
| IHT | 0.2625 | 0.0153 | 0.0288 | 0.4381 | **0.9036** | 0.3241 |
| | (0.1640) | (0.0156) | (0.0285) | (0.2543) | (0.0152) | (0.0198) |
| NM | 0.0534 | 0.8918 | 0.1061 | 0.2271 | 0.5792 | 0.0709 |
| | (0.0035) | (0.0983) | (0.0072) | (0.0096) | (0.0800) | (0.0194) |
| TL | nan | 0.0000 | 0.0000 | nan | 0.8483 | 0.2837 |
| | nan | (0.0000) | (0.0000) | nan | (0.0642) | (0.0607) |
| ENN | **0.4783** | 0.0056 | 0.0106 | **0.6747** | 0.8907 | **0.3325** |
| | (0.0001) | (0.0177) | (0.0334) | (0.0000) | (0.0126) | (0.0187) |
| OSS | nan | 0.0000 | 0.0000 | nan | 0.8692 | 0.3055 |
| | nan | (0.0000) | (0.0000) | nan | (0.0351) | (0.0382) |
| NCR | 0.1818 | 0.0031 | 0.0060 | 0.2400 | 0.8707 | 0.2928 |
| | (0.3149) | (0.0097) | (0.0189) | (0.4157) | (0.0381) | (0.0518) |
| SMOTE+ENN | 0.1112 | 0.7878 | 0.2156 | 0.3298 | 0.7957 | 0.2195 |
| | (0.0181) | (0.1055) | (0.0351) | (0.0267) | (0.0568) | (0.0309) |
| SMOTE+TL | 0.1288 | 0.7485 | 0.2442 | 0.3540 | 0.8037 | 0.2224 |
| | (0.0254) | (0.1317) | (0.0436) | (0.0349) | (0.0552) | (0.0591) |
| Average | 0.1536 | 0.5398 | 0.1419 | 0.3499 | 0.8005 | 0.2297 |
| Performance | (0.0582) | (0.0742) | (0.0335) | (0.0829) | (0.0564) | (0.0451) |

Table 9 Evaluation results of data sampling combined with LSTM-FCN

| Methods | Recall | Precision | $F_1$ | $G_{mean}$ | ROCAUC | PRAUC |
|---|---|---|---|---|---|---|
| ROS | 0.1382 | 0.9077 | 0.2683 | 0.3584 | 0.8891 | 0.3247 |
| | (0.0767) | (0.0855) | (0.1428) | (0.0978) | (0.0353) | (0.0509) |
| SMOTE | 0.2313 | 0.8536 | 0.4359 | 0.4679 | 0.8915 | 0.3321 |
| | (0.0875) | (0.0970) | (0.1545) | (0.1052) | (0.0215) | (0.0382) |
| SMOTE b1 | 0.2330 | 0.8372 | 0.4332 | 0.4695 | 0.8945 | 0.3414 |
| | (0.0896) | (0.1460) | (0.1581) | (0.1054) | (0.0267) | (0.0328) |
| SMOTE b2 | 0.2345 | 0.6765 | 0.3923 | 0.4708 | 0.8636 | 0.2943 |
| | (0.0871) | (0.1736) | (0.1076) | (0.0925) | (0.0190) | (0.0365) |
| SMOTE SVM | 0.2522 | 0.7944 | 0.4550 | 0.4908 | 0.8992 | 0.3449 |
|  | (0.0843) | (0.1507) | (0.1323) | (0.0925) | (0.0150) | (0.0241) |
| ADASYN | 0.1768 | 0.4765 | 0.2411 | 0.4028 | 0.7661 | 0.1699 |
|  | (0.0956) | (0.2201) | (0.0414) | (0.0970) | (0.0350) | (0.0334) |
| RUS | 0.1267 | 0.9250 | 0.2477 | 0.3433 | 0.8988 | 0.3303 |
|  | (0.0700) | (0.0602) | (0.1318) | (0.0937) | (0.0078) | (0.0232) |
| IHT | 0.2014 | 0.9311 | 0.3958 | 0.4424 | **0.9137** | 0.3472 |
|  | (0.0557) | (0.020) | (0.1079) | (0.0720) | (0.0120) | (0.0202) |
| NM | 0.0508 | **1.0000** | 0.1016 | 0.2254 | 0.6267 | 0.1455 |
|  | (0.0000) | (0.0000) | (0.0000) | (0.0000) | (0.0602) | (0.1351) |
| TL | 0.3359 | 0.5597 | 0.4891 | 0.5688 | 0.8994 | **0.3515** |
|  | (0.0737) | (0.1957) | (0.0579) | 0.0634 | (0.0104) | (0.0135) |
| ENN | **0.3441** | 0.5000 | 0.4592 | **0.5755** | 0.8877 | 0.3395 |
|  | (0.0688) | (0.2172) | (0.0749) | (0.0567) | (0.0190) | (0.0194) |
| OSS | 0.3325 | 0.5224 | 0.4614 | 0.5651 | 0.8950 | 0.3392 |
|  | (0.0776) | (0.1882) | (0.0761) | (0.0656) | (0.0151) | (0.0182) |
| NCR | 0.3158 | 0.6051 | **0.4984** | 0.5532 | 0.8791 | 0.3346 |
|  | (0.0590) | (0.1612) | (0.0664) | (0.0523) | (0.0173) | (0.0216) |
| SMOTE+ENN | 0.1825 | 0.8995 | 0.3471 | 0.4078 | 0.9090 | 0.3575 |
|  | (0.1047) | (0.1043) | (0.1887) | (0.1270) | (0.0097) | (0.0317) |
| SMOTE+TL | 0.3158 | 0.6051 | 0.4984 | 0.5532 | 0.8791 | 0.3346 |
|  | (0.0590) | (0.1612) | (0.0664) | (0.0523) | (0.0173) | (0.0216) |
| Average | 0.2314 | 0.7396 | 0.3816 | 0.4597 | 0.8662 | 0.3125 |
| Performance | (0.0726) | (0.1325) | (0.1005) | (0.0782) | (0.0214) | (0.0347) |

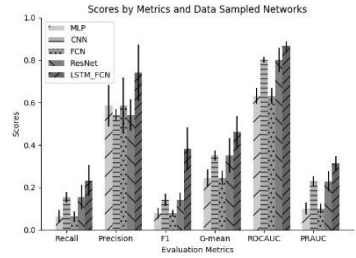

**Figure 6.** Average Performance Comparison of Data Sampled Networks

*3.6. Imbalanced time-series classification with proposed method*

In this subsection, the proposed cost-sensitive learning strategy was utilized to modify those five single neural network classifiers. The cost-sensitive networks were validated and the averages of their percentages are presented in Table 10. The best result and relevant standard deviation on each metrics are emphasized in bold-face and shown in parentheses respectively. **Figure 7** illustrates the evaluation metrics comparison of the five cost-sensitive neural networks. Results of all experiments (single classifiers, data sampled classifiers and cost-sensitive classifiers) are demonstrated in **Figures 8** through **12**.

As it is shown in Table 10, the cost-sensitive networks comprehensively could perform well and there are no zero values or invalid measures anymore. Especially, in the term of ROCAUC metrics, cost-sensitive CNN, cost-sensitive ResNet and cost-sensitive LSTM-FCN achieve high scores. As demonstrated in Figure 10, cost-sensitive

ResNet performs best except its second rank in the term of Precision metric. The comparisons are plotted in Figures 8 through 12. It is clearly observed that the proposed strategy performs better than applying single networks and data sampled networks on imbalanced time-series classification. In details, the performance of cost-sensitive CNN and cost-sensitive ResNet beat the corresponding of other methods. Cost-sensitive MLP, cost-sensitive FCN and cost-sensitive LSTM-FCN win out at least in three metrics.

Table 10 Evaluation results of five cost-sensitive networks

| Methods | Recall | Precision | $F_1$ | $G_{mean}$ | ROCAUC | PRAUC |
|---|---|---|---|---|---|---|
| MLP | 0.1288 | 0.4459 | 0.1998 | 0.3503 | 0.7053 | 0.1210 |
|  | (0.0235) | (0.1051) | (0.0385) | (0.0330) | (0.0499) | (0.0346) |
| MDCNN | 0.2930 | 0.8934 | 0.4413 | 0.5374 | 0.9082 | **0.3629** |
|  | (0.0120) | (0.0173) | (0.0141) | (0.0111) | (0.0048) | (0.0208) |
| FCN | 0.2854 | 0.2658 | 0.2752 | 0.5196 | 0.7031 | 0.1504 |
|  | (0.0641) | (0.2048) | (0.0977) | (0.0595) | (0.1000) | (0.0625) |
| ResNet | **0.3220** | 0.7730 | **0.4546** | **0.5625** | **0.9161** | 0.3457 |
|  | (0.0150) | (0.0943) | (0.0259) | (0.0121) | (0.0064) | (0.0189) |
| LSTM-FCN | 0.2199 | **0.9051** | 0.3538 | 0.4525 | 0.9035 | 0.3439 |
|  | (0.1038) | (0.0878) | (0.0951) | (0.1130) | (0.0111) | (0.0219) |

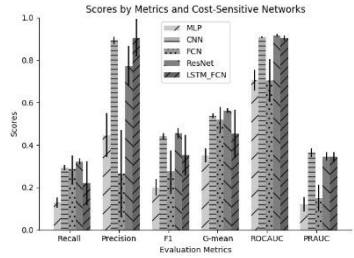

**Figure 7.** Metrics Comparison of Five Cost-sensitive Networks

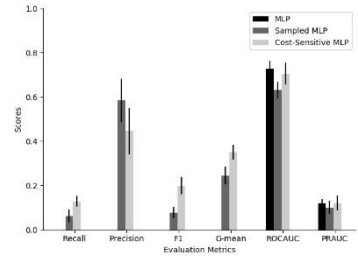

**Figure 8.** Metrics Comparison of MLP, Data Sampled MLP and Cost-sensitive MLP

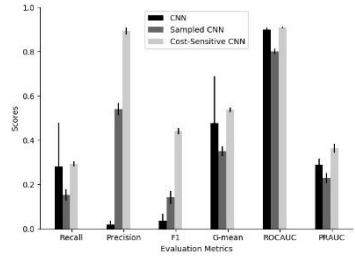

**Figure 9.** Metrics Comparison of CNN, Data Sampled CNN and Cost-sensitive CNN

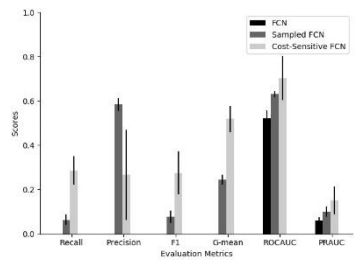

**Figure 10.** Metrics Comparison of FCN, Data Sampled FCN and Cost-sensitive FCN

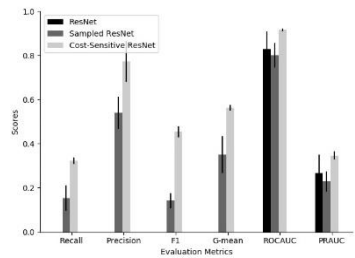

**Figure 11.** Metrics Comparison of ResNet, Data Sampled ResNet and Cost-sensitive ResNet

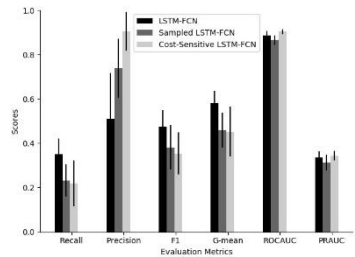

**Figure 12.** Metrics Comparison of LSTM-FCN, Data Sampled LSTM-FCN and Cost-sensitive LSTM-FCN

## 4. Discussion

### 4.1. Deep learning for TSC

In [22], the authors proved that their multi-channels deep CNN (MC-DCNN) could be used for TSC tasks. Their statement was endorsed in this paper. Considering its ability for capturing time shift features automatically, MDCNN was applied with different parameters in this paper. Additionally, some popular CNN based structures [20, 31] were also explored. But be different with them, the goal of this work was to apply deep learning on ITSC. Unfortunately, experimental results show that single neural network classifiers could not be qualified for this task due to the under-representation of minority time-series samples.

*4.2. Data sampling*

Several data samplers were applied with temporal convolutional networks in this paper. Like [30] claimed, some similar conclusions were found. The CIP issues indeed degraded the performance of deep learning models for classification and data sampling could alleviate it. They also stated that over-sampling does not necessarily cause overfitting of CNNs. However, it was uncertain to judge which sampler could win out in this paper. The experimental results illustrate that temporal CNN performs worse than single CNN classifier at least in four metrics. The difference might come from the different datasets: they validated their statement with graphic dataset while this work used temporal sequences one. Thus this work could not totally agree with their last statement and this point need to be further studied in the future work.

*4.3. Cost-sensitive Strategy*

This paper proposed an adaptive cost-sensitive learning strategy and it is appropriated for convolutional neural networks. The modified networks could be effectively applied to address ITSC problems. Although data sampling is the most direct approach to deal with ITSC issues, they would change the original distribution of raw datasets. By doing this, some drawbacks like over-fitting, useful information discarding and time-consuming might be introduced. The proposed method could avoid the above problems. Moreover, it can be automatically optimized. The proposed method can be used to classify large volume, high dimensional, imbalanced time-series. [12] and [28] did similar work. Both of them applied adaptive learning CNN for imbalanced classification and they considered overall imbalanced ratio of training set. However, the proposed strategy is different with them. The local imbalanced ratio was also taken into account in this work. It was represented by the imbalanced distribution of minibathes. In fact, their strategy can be seen as one particular case of the proposed cost-sensitive method where the distributions of minibatches are fixed to be equal with overall imbalanced ratio.

*4.4. Limitation and Future Work*

The limitation of this work was the lack of using benchmark dataset. It is because the most used benchmark dataset is UCR and it is a univariate and evenly dataset. Validation on UCR would not obey the goal of this work. Therefore, future work need conducting more experiments on appropriate time-series benchmark datasets. Furthermore, extension of the proposed method to multi-classification tasks is elementary, but has not been explored in this work. This simplest attempt could by alternating sigmoid with softmax.

## 5. Conclusion

In this work, one adaptive cost-sensitive learning strategy was proposed to deal with ITSC problems. It can be used to modify deep convolutional neural networks from cost-insensitive to cost-sensitive learning. The experimental results indicate some conclusion as follows. (1) Single neural network classifiers could not tackle the class imbalanced problems. (2) Data sampling approaches could be used to improve the performance of single networks but not always work. (3) The modified neural networks with proposed cost-sensitive ResNet. (4) The proposed cost-sensitive strategy is appropriated for convolutional neural networks and could be effectively applied to address the imbalanced time-series classification problems. In the further work, more experiments need to be done on appropriate time-series benchmark datasets and the proposed adaptive cost-sensitive strategy will be extended into multi-classification tasks.